\documentclass[journal]{IEEEtran}
\usepackage{graphicx}
\usepackage{amsmath,amssymb} 
\usepackage{color}
\usepackage{subfigure}
\usepackage{url}
\usepackage{multirow}
\usepackage{booktabs}
\usepackage{array}
\usepackage{bbding}

\ifCLASSINFOpdf
\else
\fi
\begin{document}
%
\title{Adversarial Spatio-Temporal Learning for\\ Video Deblurring}
%
%
%

\author{Kaihao~Zhang,
        Wenhan~Luo,
        Yiran~Zhong,
        Lin~Ma,
        Wei Liu,
        and~Hongdong~Li
\thanks{K. Zhang, Y. Zhong and H. Li are with the College of Engineering and Computer Science, the Australian National University, Canberra, ACT,  Australia. E-mail: \{kaihao.zhang@anu.edu.au; hongdong.li@anu.edu.au; yiran.zhong@anu.edu.au\}.

W. Luo, L. Ma and W. Liu are with the Tencent AI Laboratory, Shenzhen 518057, China. E-mail:\{whluo.china@gmail.com; forest.linma@gmail.com; wl2223@columbia.edu.\} }}

\maketitle

\begin{abstract}
Camera shake or target movement often leads to undesired blur effects in videos captured by a hand-held camera. Despite significant efforts having been devoted to video-deblur research, two major challenges remain: 1) how to model the spatio-temporal characteristics across both the spatial domain (i.e., image plane) and temporal domain (i.e., neighboring frames), and 2) how to restore sharp image details w.r.t. the conventionally adopted metric of pixel-wise errors. In this paper, to address the first challenge, we propose a {\em DeBLuRring Network (DBLRNet)} for spatial-temporal learning by applying a 3D convolution to both spatial and temporal domains. Our DBLRNet is able to capture jointly spatial and temporal information encoded in neighboring frames, which directly contributes to improved video deblur performance. To tackle the second challenge, we leverage the developed DBLRNet as a generator in the GAN (generative adversarial network) architecture, and employ a content loss in addition to an adversarial loss for efficient adversarial training.  The developed network, which we name as {\em DeBLuRring Generative Adversarial Network (DBLRGAN)}, is tested on two standard benchmarks and achieves the state-of-the-art performance.
\end{abstract}


%
\IEEEpeerreviewmaketitle


\begin{figure*}[tb]
\centering
\includegraphics[width=0.9\linewidth]{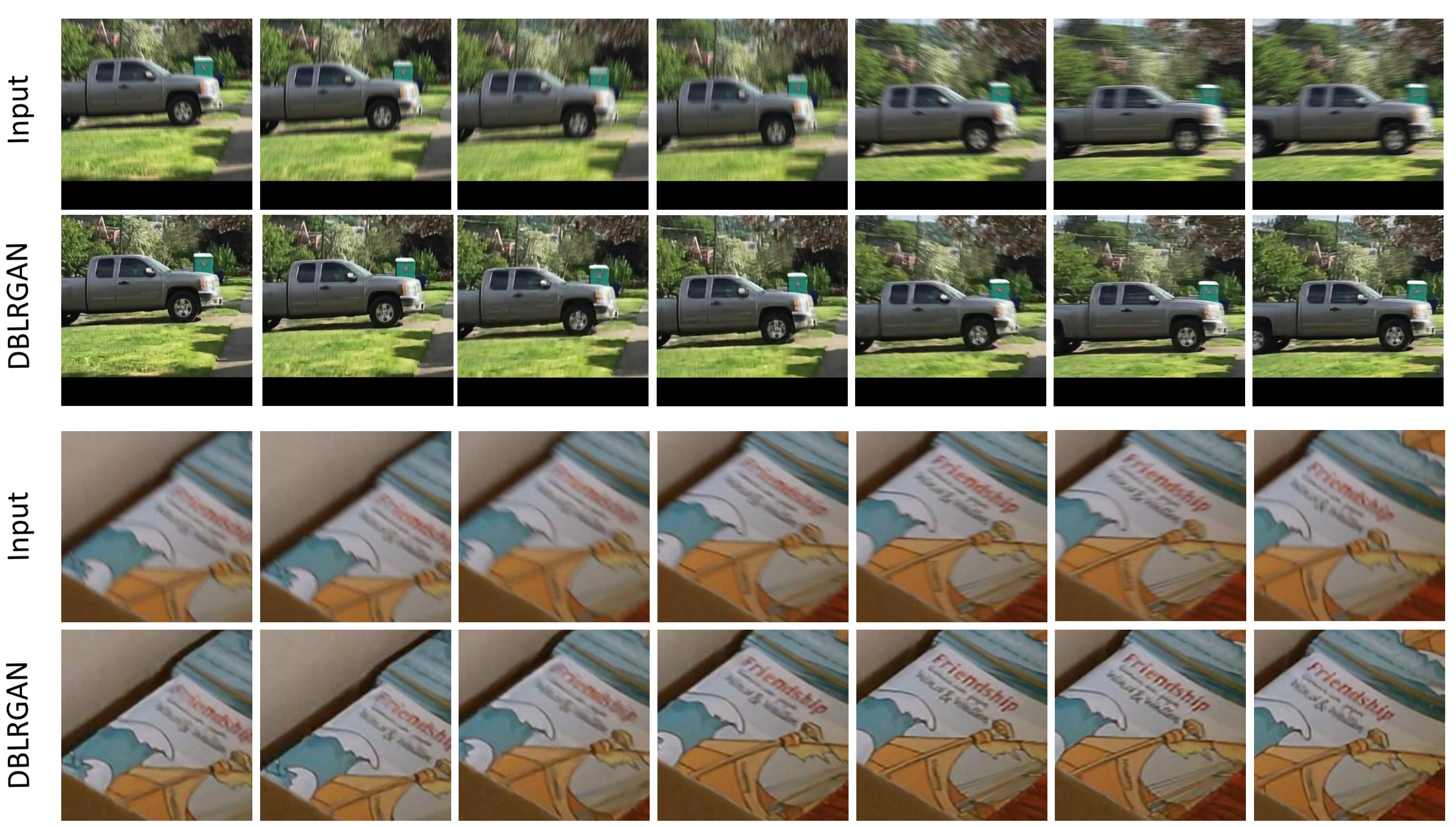}
\caption{Deblurring results of the proposed DBLRGAN on real-world video frames. The first and third rows show crops of consecutive frames from the VideoDeblurring dataset. The second and fourth rows show corresponding deblurring results of DBLRGAN. }
\label{figure1}
\end{figure*}

\section{Introduction}

Videos captured by hand-held cameras often suffer from unwanted blurs either caused by camera shake \cite{kang2007automatic}, or object movement in the scene \cite{sun2015learning,shi2015just}. The task of video deblurring aims at removing those undesired blurs and recovering sharp frames from the input video. This is an active research topic in the applied fields of computer vision and image processing.  Applications of video deblurring are found in many important fields such as 3D reconstruction \cite{seok2013dense}, SLAM \cite{lee2011simultaneous} and tracking \cite{jin2005visual}. 

In contrast to single image deblurring, video deblurring is a relatively less tapped task until recently.  And video deblurring is more challenging, partly because  it is not entirely clear about how to model and exploit the inherent temporal dynamics exhibited among continuous video frames. Moreover, the commonly adopted performance metric, namely, pixel-wise residual error, often measured by PSNR, is questionable, as it fails to capture human visual intuitions of how sharp or how realistic a restored image is  \cite{wang2003multiscale,wang2004image}.  In this paper, we plan to leverage the recent advance of the adversarial learning technique to improve the performance of video deblurring.

One key challenge for video deblurring is to find an effective way to capture spatio-temporal information existing in neighboring image frames. Deep learning based methods have recently witnessed a remarkable success in many applications including image and video denoising and deblurring. Previous deep learning methods are however primarily based on 2D convolutions, mainly for computational sake. Yet, it is not natural to use 2D convolutions to capture spatial and temporal joint information, which is essentially in a 3-D feature space. In this paper, we propose a deep neural network called DeBLuRing Network (DBLRNet), which uses 3D (volumetric) convolutional layers, as well as deep residual learning, aims to learn feature representations both across temporal frames and across image plane.  

As noted above, we argue that the conventional pixel-wise PSNR metric is insufficient for the task of image/video deblurring. To address this issue, we resort to adversarial learning, and propose DeBLuRring Generative Adversarial Network (DBLRGAN). DBLRGAN consists of a generative network and a discriminate network, where the \textit{\textbf{generative}} network is the aforementioned DBLRNet which restores sharp images, and the \textit{\textbf{discriminate}} network is a binary classification network, which tells a restored image apart from a real-world sharp image.

We introduce a training loss which consists of two terms: content loss and adversarial loss. The content loss is used to respect the pixel-wise measurement, while the adversarial loss promotes a more realistically looking (hence sharper) image.  Training DBLRGAN in an end-to-end manner, we recover sharp video frames from a blurred input video sequence, with some examples shown in Figure~\ref{figure1}.  

The contributions of this work are as follows:
\begin{itemize}
\item
We propose a model called DBLRNet, which applies  3D convolutions in a deep residual network to capture joint spatio-temporal features for video deblurring.
\item
Based on the above DBLRNet, we develop a generative adversarial network, called DBLRGAN, with both content and adversarial losses. By training it in an adversarial manner, the DBLRGAN recovers video frames which look more realistic.
\item Experiments on two standard benchmark datasets, including the VideoDeblurring dataset and the Blurred KITTI dataset, show that the proposed network DBLRNet and DBLRGAN are effective and outperforms existing methods.
\end{itemize}

\section{Related Work}

Many approaches have been proposed for image/video deblurring, which can be roughly classified into two categories: geometry-based methods and deep learning methods.

\textbf{Geometry-based methods.} Modern single-image deblurring methods iteratively estimate uniform or non-uniform blur kernels and the latent sharp image given a single blurry image \cite{gupta2010single,hirsch2011fast,hu2014joint,Jin_2017_CVPR,Srinivasan_2017_CVPR,Yan_2017_CVPR,Pan_2017_ICCV, Gong_2017_ICCV, Dong_2017_ICCV, Bahat_2017_ICCV, R._2017_ICCV, Park_2017_ICCV}. However, it is difficult for single image based methods to estimate kernel because blur is spatially varying in real world. To employ additional information, multi-image based methods \cite{ito2014blurburst,yuan2007image,petschnigg2004digital,hee2014gyro,tai2008image, Wieschollek_2017_ICCV,Ren_2017_ICCV} have been proposed to address blur, such as flash/no-flash image pairs \cite{petschnigg2004digital}, blurred/noise image pairs \cite{yuan2007image} and gyroscope information \cite{hee2014gyro}. In order to accurately estimate kernels, some methods also use optical flow \cite{hyun2015generalized} and temporal information \cite{li2010generating}.
However, most of these methods are limited by the performance of an assumed degradation model and its estimation, thus some of them are fragile and cannot handle more challenging cases. 

Some researchers attempt to use aggregation methods to alleviate blur. Law et al. \cite{law2006lucky} propose a lucky image system, which constructs a final image based on the best pixels from different low quality images. Cho et al. \cite{cho2012video} use patch-based synthesis to restore blurry regions and ensure that the deblurred frames are spatially and temporally coherent. Motivated from the physiological fact, an efficient Fourier aggregation method is proposed in \cite{delbracio2015burst}, which creates a consistently registered version of neighboring frames, and then fuses these frames in the Fourier domain. 

More recently, Pan et al. \cite{pan2017simultaneous} propose to simultaneously deblur stereo videos and estimate the scene flow. In this method, motion cues from scene flow estimation, and blur information can complement each other and boost the performance. However, this kind of approaches is restricted to stereo cameras.

\begin{figure*} [ht]
	\centering
	\includegraphics[width=0.99\linewidth]{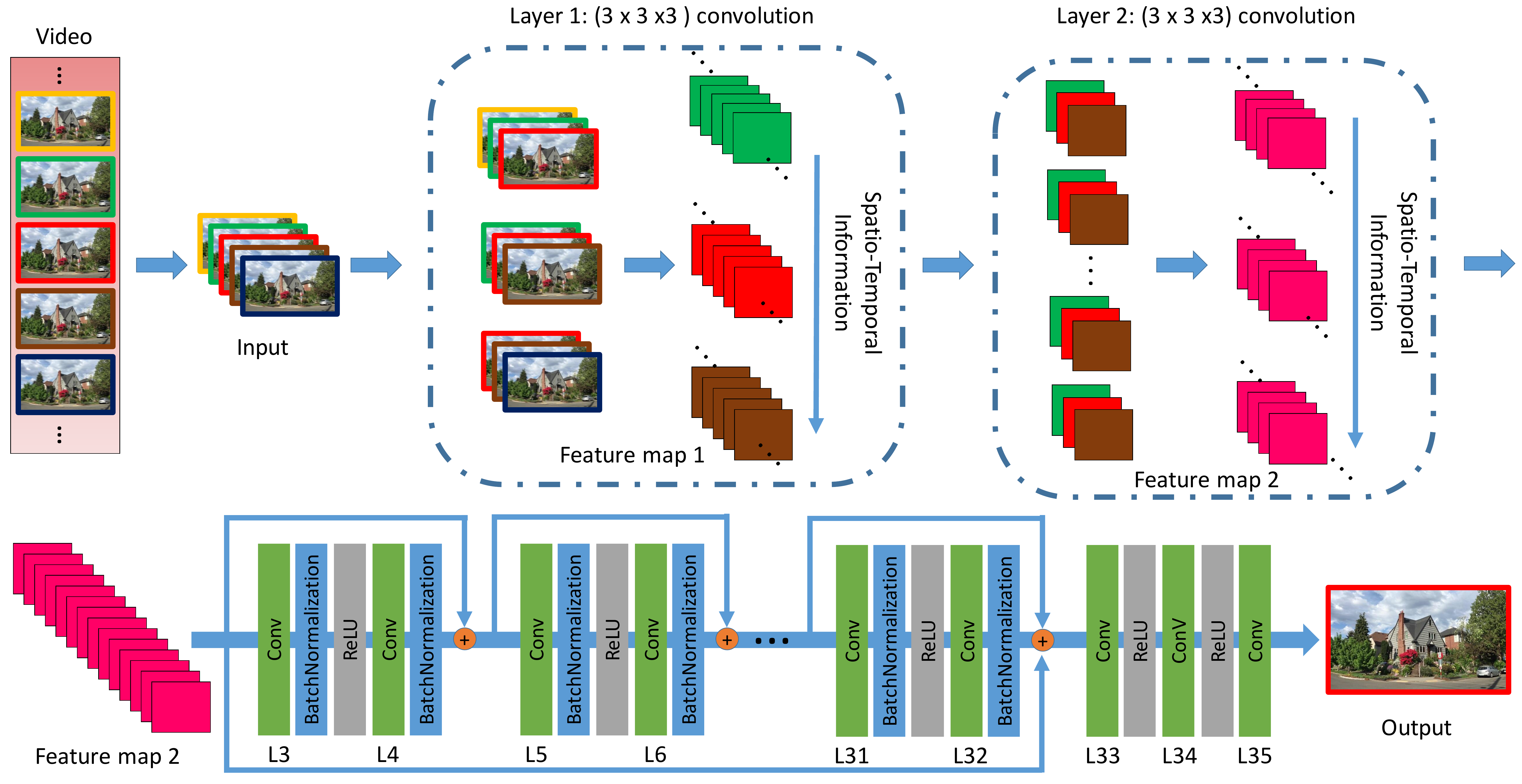}
	\caption{The proposed DBLRNet framework. The input to our network is five time-consecutive blurry frames. The output is the central deblurred frame. By performing 3D convolutions, this model learns joint spatial-temporal feature representations.}
	\label{figure2}
\end{figure*}

\textbf{Deep learning methods.} Deep learning has shown its effectiveness in many computer vision tasks, such as object detection \cite{ren2015faster}, image classification \cite{krizhevsky2012imagenet,simonyan2014very,szegedy2015going,he2016deep}, facial processing \cite{sun2013deep,sun2014deep,sun2015learning,zhang122015kinship,zhang2017facial} and multimedia analysis \cite{ji20133d,simonyan2014two,ledig2016photo}. There are also several deep learning methods that achieve encouraging results on deblurring \cite{xu2014deep,sun2015learning,nah2016deep,su2016deep,Nimisha_2017_ICCV,kim2017online,kim2017dynamic}. A non-bind deblurring method with deep Convolutional Neural Network (CNN) is proposed in \cite{xu2014deep}. This data-driven approach establishes connection between traditional optimization-based schemes and empirically-determined CNN. Sun et al. \cite{sun2015learning} predict the probabilistic distribution of motion kernels at the patch level using CNN, then use a Markov random field model to fuse the estimations into a dense field of motion kernels. Finally, a non-uniform deblurring model using patch-level prior is employed to remove motion blur. In \cite{nah2016deep}, a deep multi-scale CNN is proposed for image deblurring. Most of these methods aim at image deblurring, so they do not need to consider temporal information implied in videos.

For video deblurring, the method closest to our approach is DBN \cite{su2016deep}, which proposes a CNN model to process information across frames. Neighboring frames are stacked along RGB channels and then fed into the proposed model to recover the central frame of them. This method considers multiple frames together and thus achieves comparable performance with state-of-the-art methods.

However, this method employs 2D convolutions, which do not operate in the time axis (corresponding to temporal information). By doing so, temporal information is transformed into spatial information in their setting, thus limited temporal information is preserved. Meanwhile, this method (as most of the existing methods) trains the model to maximize the pixel fidelity, which cannot ensure that the recovered images look realistic sharp. Our proposed method, on the contrary, learns spatio-temporal features by 3D convolutions, and integrates the 3D deblurring network into a generative adversarial network to achieve photo-realistic results.

Even our proposed method takes neighboring frames as inputs, we call it as video deblurring method for two reasons. Firstly, previous work like DBN \cite{su2016deep}, is called as video deblurring method. DBN also takes five neighboring frames as input to generate the middle sharp frame. Secondly, this method can be applied to tackle video deblurring task in the real world. Specially, videos can be regarded as multiple consecutive frames. When videos are input into our model, the proposed DBLRNet tackles five neighboring frames as a whole to generate the deblurred middle frame based on their spatio-temporal information, and obtain the deblurred videos by continuous deblurred frames finally.

\section{Our Model}

\textbf{Overview.} In this section, we first introduce our DBLRNet, and then present the proposed network DBLRGAN which is on the basis of DBLRNet. Finally we detail the two loss functions (content and adversarial losses) which are used in the training stage. Both the DBLRNet and DBLRGAN are end-to-end systems for video deblurring. Note that, blurry frames can be put into our proposed models without alignment. 

\subsection{DBLRNet}

\noindent In 2D CNN, convolutions are applied on 2D images or feature maps to learn features in spatial dimensions only. In case of video analysis problems, it is desirable to consider the motion variation encoded in the temporal dimension, such as multiple neighboring frames. In this paper, we propose to perform 3D convolutions \cite{ji20133d} the convolution stages of deep residual networks to learn feature representations from both spatial and temporal dimensions for video deblurring. We operate the 3D convolution via convolving 3D kernels/filters with the cube constructed from multiple neighboring frames. By doing so, the feature maps in the convolution layers can capture the dynamic variations, which is helpful to model the blur evolution and further recover sharp frames.

Formally, the 3D convolution operation is formulated as:

\begin{equation}
\footnotesize
V_{ij}^{xyz} = \sigma (\sum\limits_m {\sum\limits_{p = 0}^{{P_i} - 1} {\sum\limits_{q = 0}^{{Q_i} - 1} {\sum\limits_{r = 0}^{{R_i} - 1} {V_{(i - 1)m}^{(x + p)(y + q)(z + r)} \cdot g_{ijm}^{pqr}}  + {b_{ij}}} } } ) \, ,
\end{equation}
where $V_{ij}^{xyz}$ is the value at position $(x,y,z)$ in the $j$-th feature map of the $i$-th layer, (${P_i}$, ${Q_i}$, ${R_i}$) is the size of 3D convolution kernel. ${Q_i}$ responds to the temporal dimension. $g_{ijm}^{pqr}$ is the $(p,q,r)$-th value of the kernel connected to the $m$-th feature map from the $(i-1)$-th layer. $\sigma\left(\cdot\right)$ is the ReLU nonlinearity activation function, which is shown to lead to better performance in various computer vision tasks than other activation functions, e.g. Sigmoid and Tanh.

Defining 3D convolution, we propose a model called DBLRNet, which is shown in Figure~\ref{figure2}. DBLRNet is composed of two $3 \times 3 \times 3$ convolutional layers, several residual blocks \cite{he2016deep}, each containing two convolution layers, and another five convolutional layers. This architecture is designed inspired by the Fully Convolutional Neural Network (FCNN) \cite{long2015fully}, which is originally proposed for semantic segmentation. Different from FCNN and DBN \cite{su2016deep}, spatial size of feature maps in our model keeps constant. Namely, there is not any down-sampling operation nor up-sampling operation in our DBLRNet. The detailed configurations of DBLRNet is given in Table \ref{table1}.

\begin{figure*}
	\centering
	\includegraphics[width=1\linewidth]{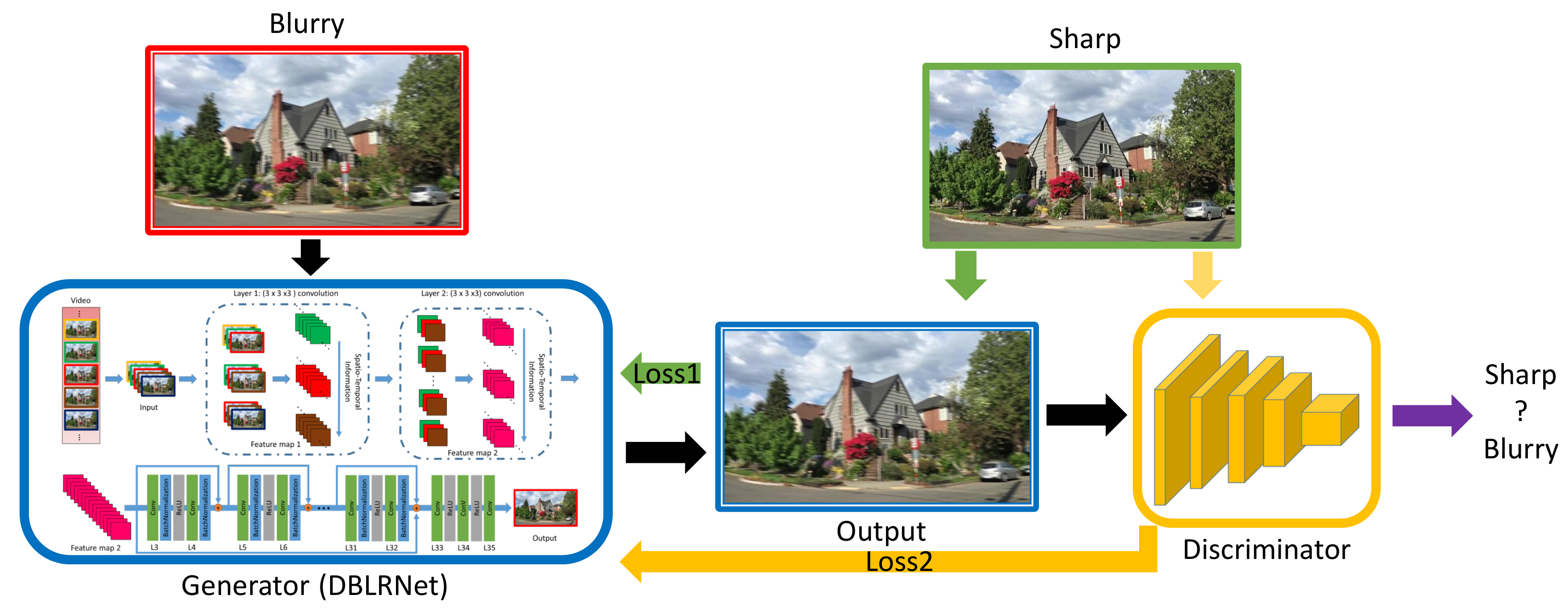}
    \caption{The DBLRGAN framework for video deblurring. The architecture consists of a Generator and a Discriminator. The Generator is our proposed DBLRNet, while the Discriminator is a VGG-like CNN net.}
	\label{figure3}
\end{figure*}

\begin{table}
  \centering
  \scriptsize
  \caption{Configurations of the proposed DBLRNet. It is composed of two convolutional layers (L1 and L2), 14 residual blocks, two convolutional layers (L31 and L32)  without skip connection, and three additional convolutional layers (L33, L34 and L35). Each residual block contains two convolutional layers, which are indicated by L(X) and L(X+1) in the Table, where ``X'' equals 3, 5, 7, 9, 11, 13, 15, 17, 19, 21, 23, 25, 27 and 29 respectively for these residual blocks.}
    \begin{tabular}{cccccc}
    \toprule
    layers & Kernel size & output channels & operations & skip connection\\
    \midrule
     L1     & $3 \times 3 \times 3$  & 16 & ReLU & - \\
     L2       & $3 \times 3 \times 3$  & 64 & ReLU & L4, L32 \\    
     \midrule
     L3     & $3 \times 3 \times 1$  & 64 & BN + ReLU & - \\
     L4     & $3 \times 3 \times 1$  & 64 & BN        & L6 \\     
     L5     & $3 \times 3 \times 1$  & 64 & BN + ReLU & - \\
     L6     & $3 \times 3 \times 1$  & 64 & BN        & L8 \\        
     L7     & $3 \times 3 \times 1$  & 64 & BN + ReLU & - \\
     L8     & $3 \times 3 \times 1$  & 64 & BN        & L10 \\ 
     L9     & $3 \times 3 \times 1$  & 64 & BN + ReLU & - \\
     L10    & $3 \times 3 \times 1$  & 64 & BN        & L12 \\ 
     L11    & $3 \times 3 \times 1$  & 64 & BN + ReLU & - \\
     L12    & $3 \times 3 \times 1$  & 64 & BN        & L14 \\
     L13    & $3 \times 3 \times 1$  & 64 & BN + ReLU & - \\
     L14    & $3 \times 3 \times 1$  & 64 & BN        & L16 \\
     L15    & $3 \times 3 \times 1$  & 64 & BN + ReLU & - \\
     L16    & $3 \times 3 \times 1$  & 64 & BN        & L18 \\ 
     L17    & $3 \times 3 \times 1$  & 64 & BN + ReLU & - \\
     L18    & $3 \times 3 \times 1$  & 64 & BN        & L20 \\ 
     L19    & $3 \times 3 \times 1$  & 64 & BN + ReLU & - \\
     L20    & $3 \times 3 \times 1$  & 64 & BN        & L22 \\ 
     L21    & $3 \times 3 \times 1$  & 64 & BN + ReLU & - \\
     L22    & $3 \times 3 \times 1$  & 64 & BN        & L24 \\ 
     L23    & $3 \times 3 \times 1$  & 64 & BN + ReLU & - \\
     L24    & $3 \times 3 \times 1$  & 64 & BN        & L26 \\   
     L25    & $3 \times 3 \times 1$  & 64 & BN + ReLU & - \\
     L26    & $3 \times 3 \times 1$  & 64 & BN        & L28 \\ 
     L29    & $3 \times 3 \times 1$  & 64 & BN + ReLU & - \\
     L30    & $3 \times 3 \times 1$  & 64 & BN        & L32 \\ 
     \midrule
     L31    & $3 \times 3 \times 1$  & 64 & BN + ReLU & - \\
     L32    & $3 \times 3 \times 1$  & 64 & BN        & - \\
 	\midrule
     L33    & $3 \times 3 \times 1$  & 256&     ReLU & - \\
     L34    & $3 \times 3 \times 1$  & 256&     ReLU & - \\
     L35    & $3 \times 3 \times 1$  & 1  &    -     & - \\
    \bottomrule
    \end{tabular}
  \label{table1}
\end{table}

As Figure~\ref{figure2} shows, the input to DBLRNet is five consecutive frames. Note that we does not conduct deblurring in the original RGB space. Alternatively, we conduct deblurring on basis of gray-scale images. Specifically, the RGB space is transformed to the YCbCr space, and the Y channel is adopted as input since the illumination is the most salient one. We perform 3D convolutions with kernel size of $3 \times 3 \times 3$ ($3 \times 3$ is the spatial size and the last 3 is for the temporal dimension) in the first and second convolutional layers. To be more specific, in layer 1, three groups of consecutive frames are convolved with a set of 3D kernels respectively, resulting in three groups of feature maps. These three groups of feature maps are convolved with 3D filters again to obtain higher-level feature maps. In the following layers, the size of convolution kernels is $3 \times 3 \times 1$ due to the decrease of temporal dimensions. The stride and padding are set to 1 in every layer. The output of DBLRNet is the deblurred central frame. We transform the gray-scale output back to colorful images with the original Cb and Cr channels.

\begin{table}
\centering
   \scriptsize
   \caption{Configurations of our D model in DBLRGAN. BN means batch normalization and ReLU represents the activation function.}
    \begin{tabular}{c|cccccc}
    \toprule
     Layers & 1-2 & 3-5 & 6-9 & 10-14 & 15-16 & 17 \\ 
     \midrule
     kernel & 3 x 3 & 3 x 3 & 3 x 3 & 3 x 3 & FC & FC \\ 
     channels & 64 & 128 & 256 & 512 & 4096 & 2 \\
     BN & BN & BN & BN & BN & - & - \\
     ReLU & ReLU & ReLU & ReLU & ReLU & - & - \\
    \bottomrule
    \end{tabular}
  \label{table2}
\end{table}

\subsection{DBLRGAN}
GAN is proposed to train generative parametric models by \cite{goodfellow2014generative}. It consists of two networks: a generator network G and a discriminator network D. The goal of G is to generate samples, trying to fool D, while D is trained to distinguish generated samples from real samples. Inspired by the adversarial training strategy, we propose a model called DeBLuRring Generative Adversarial Network (DBLRGAN), which utilizes G to deblur images and D to discriminate deblurred images and real-world sharp images. Ideally, the discriminator can be fooled if the generator outputs sharp enough image.

Following the formulation in \cite{goodfellow2014generative}, solving the deblurring problem in the generative adversarial framework leads to the following min-max optimization problem:

\begin{equation}
\begin{array}{l}
\mathop {\min }\limits_G \mathop {\max }\limits_D V(G,D) = {{\rm E}_{h \sim {p_{train(h)}}}}[\log (D(h))] + \\
\ \ \ \ \ \ \ \ \ \ \ \ \ \ \ \ \ \ \ \ \ \ \ \ \ \ \ \ \ \ \ \ \ \ \ \ {{\rm E}_{\hat{h} \sim {p_{G(\hat{h})}}}}[\log (1 - D(G(\hat{h})))]\, ,
\end{array}
\end{equation}
where $h$ indicates a sample from real-world sharp frames and $\hat{h}$ represents a blurry sample. G is trained to fool D into misclassifying the generated frames, while D is trained to distinguish deblurred frames from real-world sharp frames. G and D models are trained alternately, and our ultimate goal is to train a model G that recovers sharp frames given blurry frames.

As shown in Figure~\ref{figure3}, we use the proposed DBLRNet (Figure~\ref{figure2} and Table~\ref{table1}) as our G model, and build a CNN model as our D model, following the architectural guidelines proposed by Radford et al. \cite{radford2016unsupervised}. This D model is similar to the VGG network \cite{simonyan2014very}. It contains 14 convolutional layers. From bottom to top, the number of channels of the convolutional kernels increases from 64 to 512. Finally, this network is trained via a two-way soft-max classifier at the top layer to distinguish real sharp frames from deblurred ones. For more detailed configurations, please refer to Table \ref{table2}.

\subsection{Loss Functions}

In our work, we use two types of loss functions to train DBLRGAN.

\textbf{Content Loss.} The Mean Square Error (MSE) loss is widely used in optimization objective for video deblurring in many existing methods. Based on MSE, our content loss function is defined as:

\begin{equation}
{\mathcal{L}_{content}} = \frac{1}{{WH}}\sum\limits_{x = 1}^W {\sum\limits_{y = 1}^H {{{(I_{x,y}^{sharp} - G(I^{blurry})_{x,y})}^2}} }\, ,
\end{equation}
where $W$ and $H$ are the width and height of a frame, $I_{x,y}^{sharp}$ is the value of sharp frames at location $\left(x,y\right)$, and $G(I^{blurry})_{x,y}$ corresponds to the value of deblurred frames which are generated from DBLRNet.

\textbf{Adversarial Loss.} In order to drive G to generate sharp frames similar to the real-world frames, we introduce an adversarial loss function to update models. During the training stage, parameters of DBLRNet are updated in order to fool the discriminator D. The adversarial loss function can be represented as:

\begin{equation}
{\mathcal{L}_{adversarial}} = \log (1 - D(G({I^{blurry}})))\, ,
\end{equation}
where $D(G({I^{blurry}})$ is the probability that the recovered frame is a real sharp frame.

\textbf{Balance of Different Loss Functions.} In the training stage, the loss functions are combined in a weight fusion fashion:

\begin{equation}
\mathcal{L} = \mathcal{L}_{content} + \alpha \cdot \mathcal{L}_{adversarial} \, .
\end{equation}

In order to balance the content and adversarial losses, we use a hyper-parameter  $\alpha$ to yield the final loss $\mathcal{L}$. We investigate different values of $\alpha$ from 0 to 0.1. When $\alpha=0$, only the content loss works. In this case, DBLRGAN degrades to DBLRNet. With the increase of $\alpha$, the adversarial loss plays a more and more important role. The value of $\alpha$ should be relative small, because large values of $\alpha$ can degrades the performance of our model.

\begin{figure*} [ht] 
	\centering
	\includegraphics[width=1\linewidth]{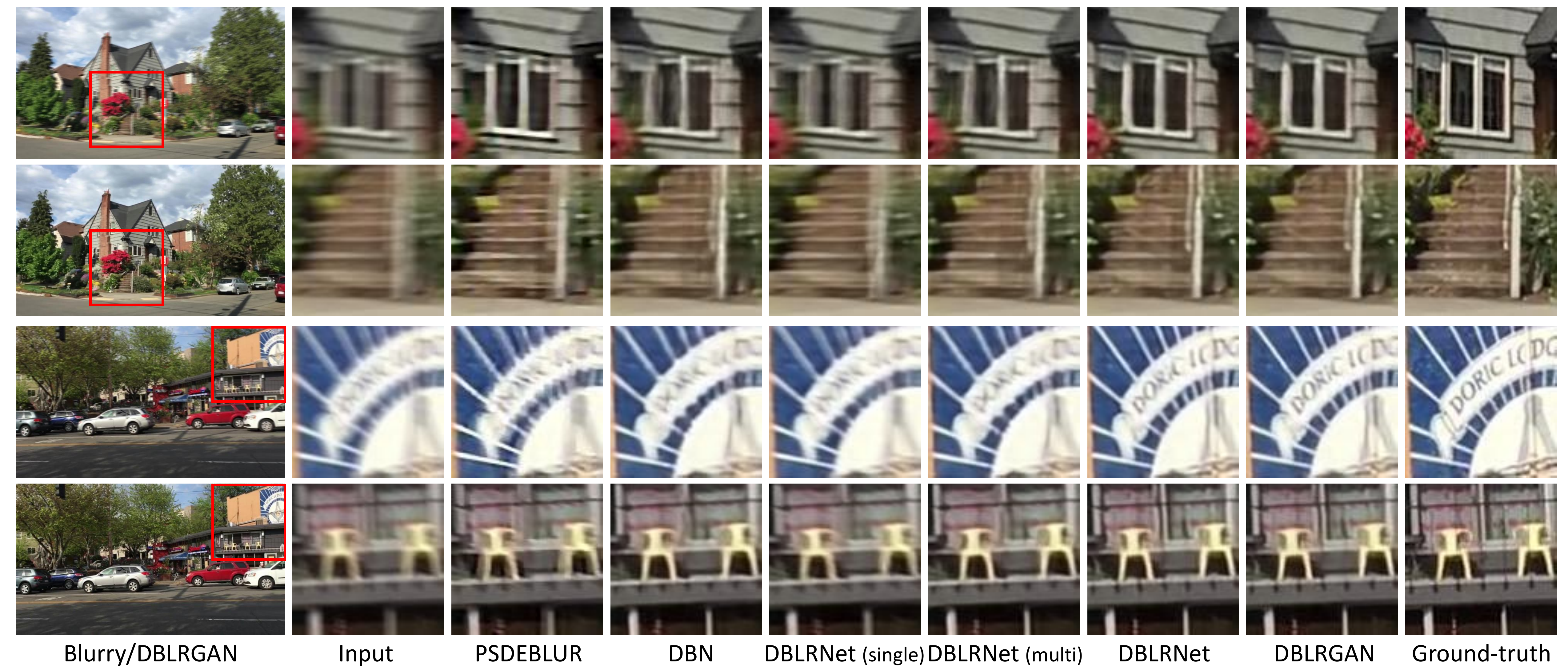}
	\caption{Exemplar results on the VideoDeblurring dataset (quantitative subset). From left to right: real blurry frame/ Output of DBLRGAN, input, PSDEBLUR, DBN \cite{su2016deep}, DBLRNet (single), DBLRNet (multi), DBLRNet, DBLRGAN and ground-truth. All results are obtained without alignment. Best viewed in color.}
	\label{figure4}
\end{figure*}

\section{Experimental Results}
\noindent In this section, we conduct experiments to demonstrate the effectiveness of the proposed DBLRNet and DBLRGAN on the task of video deblurring.

\subsection{Datasets}

\textbf{VideoDeblurring Dataset.} Su et al. build a benchmark which contains videos captured by various kinds of devices such as iPhone 6s, GoPro Hero 4 and Nexus 5x, and each video includes about 100 frames of size 1280 $\times$ 720 \cite{su2016deep}. This benchmark consists of two sub datasets: quantitative and qualitative ones. The quantitative subset contains 6708 blurry frames and their corresponding ground-truth sharp frames from 71 videos. The qualitative subset includes 22 scenes, most of which contain more than 100 images. Note that there is not ground truth for the qualitative subset, thus we can only conduct qualitative experiments on this subset. We split the quantitative subset into 61 training videos and 10 testing videos, which is the same setting as the previous method \cite{su2016deep}. Besides quantitative experiments on the 10 testing videos, we additionally test our models on the qualitative subset.

\begin{table*} [htb] 
  \centering
  \caption{Performance comparisons in terms PSNR with PSDEBLUR, WFA \cite{delbracio2015burst}, DBN (single), DBN (noalign), DBN(flow)~\cite{su2016deep}, DBLRNet (single) and DBLRNet (multi) on the VideoDeblurring dataset. The best results are shown in bold, and the second best are underlined. All results of DBLRNet and DBLRGAN are obtained without aligning.}
    \begin{tabular}{c|cccccccccc|c}
    \toprule
  Methods   & 1 & 2 & 3 & 4 & 5 & 6 & 7 & 8 & 9 & 10 & Average (PSNR) \\
    \midrule
    INPUT   & 24.14 & 30.52  & 28.38  & 27.31  & 22.60  & 29.31  & 27.74  & 23.86 & 30.59 & 26.98 & 27.14 \\
    PSDEBLUR   & 24.42  & 28.77  & 25.15  & 27.77  & 22.02  & 25.74  & 26.11  & 19.71 & 26.48 & 24.62 & 25.08  \\
    WFA  & 25.89  & 32.33  & 28.97  & 28.36  & 23.99  & 31.09  & 28.58  & 24.78 & 31.30 & 28.20 & 28.35 \\
    DBN (single)   & 25.75  & 31.15  & 29.30  & 28.38  & 23.63  & 30.70  & 29.23  & 25.62 & 31.92 & 28.06 & 28.37 \\
    DBN (noalign)  & 27.83  & 33.11  & 31.29  & 29.73  & 25.12  & 32.52  & 30.80  & 27.28 & 33.32 & 29.51 & 30.05 \\
    DBN (flow)  & 28.31  & 33.14  & 30.92  & 29.99  & 25.58  & 32.39  & 30.56  & 27.15 & 32.95 & 29.53 & 30.05 \\
  \hline
    DBLRNet (single)  & 28.68  & 29.40  & 35.11  & 32.25  & 24.94  & 30.77  &  29.81 & 25.67 & 33.14 & 30.06 & 29.98  \\
    DBLRNet (multi)  & 30.40  & 32.17  & 36.68  & 33.38  & 26.20  &  32.20 & 30.71  & 26.71 & 36.50 & 30.65 & 31.56 \\
    DBLRNet   & \underline{31.96}  & \underline{34.31}  & \textbf{37.86}  & \textbf{35.21}  & \underline{27.23}  & \underline{33.63}  & \underline{32.32}  & \underline{27.84} & \underline{38.23} & \underline{31.83} & \underline{33.04} \\
   \hline
    \textbf{DBLRGAN} & \textbf{32.32}  &\textbf{34.51}  & \underline{37.63}  & \underline{35.18}  & \textbf{27.42}  & \textbf{33.81}  & \textbf{32.43}  & \textbf{28.18} & \textbf{38.32} & \textbf{32.06} & \textbf{33.19} \\
    \bottomrule
    \end{tabular}
  \label{table3}
\end{table*}

\textbf{Blurred KITTI Dataset.} Geiger et al. develop a dataset called KITTI by using their autonomous driving platform \cite{geiger2013vision}. The KITTI dataset consists of several subsets for various kinds of tasks, such as stereo matching, optical flow estimation, visual odometry, 3D object detection and tracking. Based on the stereo 2015 dataset in the KITTI dataset, Pan et al. create a synthetic Blurred KITTI dataset \cite{pan2017simultaneous}, which contains 199 scenes. Each of the scenes includes 3 images captured by a left camera and 3 images captured by a right camera. It is worthy noting that, the KITTI data set is not used when training our models. Namely, this dataset is utilized only for testing.

\subsection{Implementation Details and Parameters}
When training DBLRNet, we use Gaussian distribution with zero mean and a standard deviation of 0.01 to initialize weights. In each iteration, we update all the weights after learning a mini-batch of size 4. To augment the training set, we crop a $128 \times 128$ patch at any location of an image ($1280 \times 720$). In this way, there are at least 712193 possible samples per one frame on the dataset \cite{su2016deep}, which greatly increases the number of training samples. In addition, we also randomly flip frames in the training stage. The DBLRNet is trained with a learning rate of $10^{-4}$, based on the content loss only. We also decrease the learning rate to $10^{-5}$ when the training loss does not decrease (usually after about 1.5 x $10^{5}$ iterations), for the sake of additional performance improvement.

In DBLRGAN, we set the hyper parameter $\alpha$ as 0.0002 when we conduct experiments as empirically this value achieves the best performance. It has a better PSNR value due to three reasons. Firstly, when training DBLRGAN, we directly place DBLRNet as our generator and fine-tune our DBLRGAN. Thus, the DBLRGAN has a high PSNR value like DBLRNet at the beginning. Secondly, the loss functions of DBLRGAN are combined in a weight fusion fashion. We set the hyper parameter $\alpha$ as 0.0002 when we conduct experiments. This is a very small value, which forces the content loss to have an overwhelming superiority over the adversarial loss on PSNR value during the training stage. Thirdly, the learning rate is set as $10^{-5}$, so the PSNR value does not have severe changes. We early stop training our DBLRGAN before the PSNR start to drop.

\begin{figure*} [ht]
	\centering
	\includegraphics[width=0.95\linewidth]{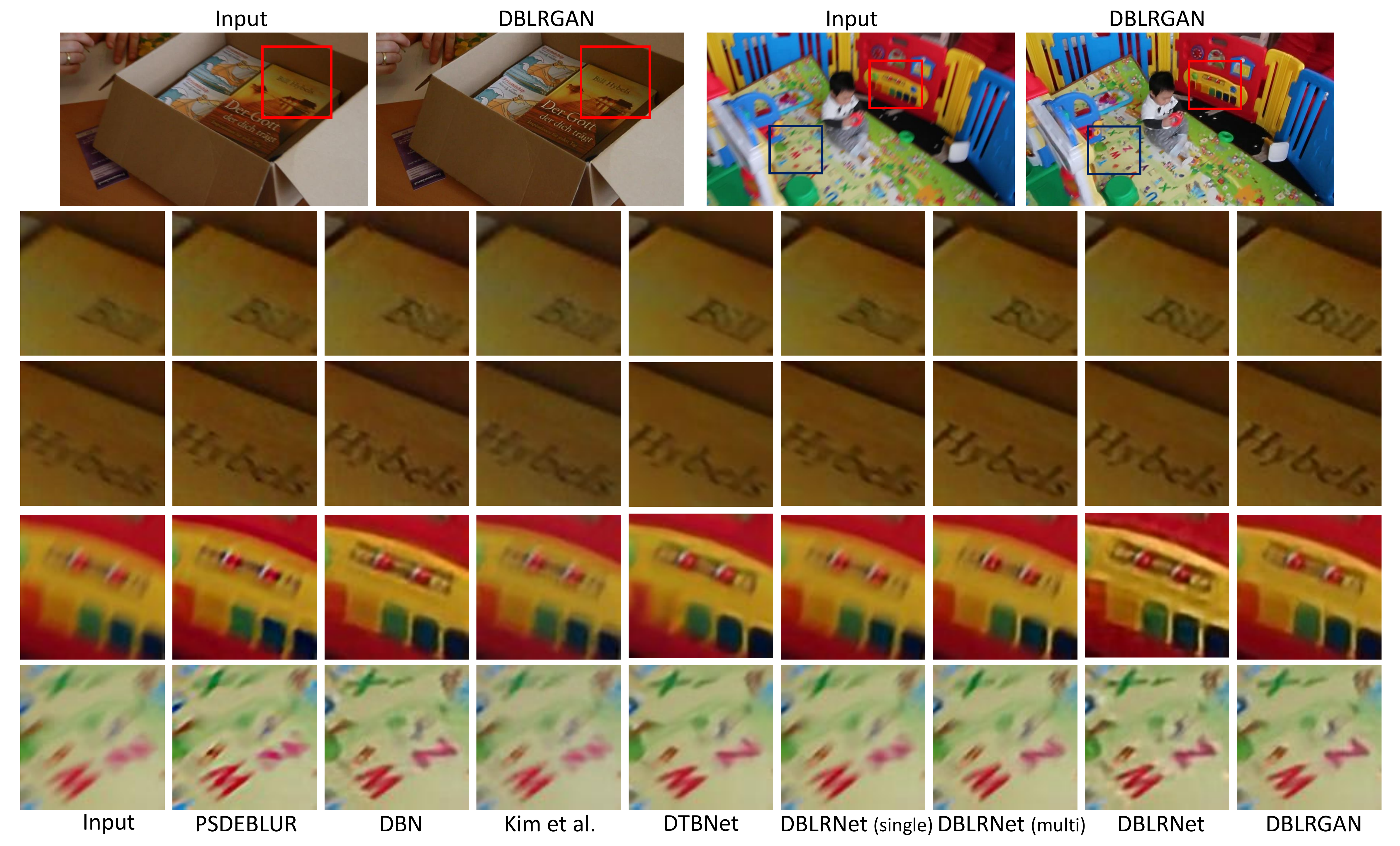}
	\caption{ Exemplar results on the VideoDeblurring dataset (qualitative subset). From left to right: real blurry frame/Output of GBLRGAN, input, PSDEBLUR, DBN \cite{su2016deep}, Kim et al. \cite{hyun2015generalized}, DTBNet~\cite{kim2017online}, DBLRNet (single), DBLRNet (multi), DBLRNet and DBLRGAN. All results are attained without alignment. Best viewed in color.}
	\label{figure5}
\end{figure*}

\subsection{Effectiveness of DBLRNet}
The proposed DBLRNet has the advantage of learning spatio-temporal feature representations. In order to verify the effectiveness of DBLRNet, we develop another two similar neural networks: DBLRNet (single) and DBLRNet (multi). These two models have the same network architectures as the original DBLRNet while there are two differences between them and the original DBLRNet. The first difference is the input. The input of DBLRNet (single) is one single frame, while the input of DBLRNet (multi) and DBLRNet is a stack of five neighboring frames. The second difference is that, in both DBLRNet (single) and DBLRNet (multi), all the convolution operations are 2D convolution operations. 

Table \ref{table3} and \ref{table4} show the PSNR values of DBLRNet (single), DBLRNet (multi) and DBLRNet on the VideoDeblurring dataset and the Blurred KITTI dataset, respectively. Compared with DBLRNet (single), DBLRNet (multi) achieves approximately 3\% $\sim$ 5\% improvement of PSNR values, which shows that stacking multiple neighboring frames is useful to learn temporal features for video deblurring even in case of 2D convolution. Comparing DBLRNet with DBLRNet (multi), there are additionally 1\% $\sim$ 5\% improvement in terms of PSNR. We suspect that the improvement results from the power of spatio-temporal feature representations learned by 3D convolution. Conducting these two kinds of comparisons, the effectiveness of DBLRNet has been verified.

\subsection{Effectiveness of DBLRGAN}
In this section, we investigate the performance of the proposed DBLRGAN. Table \ref{table3} and \ref{table4} show the quantitative results on the VideoDeblurring and Blurred KITTI dataset, respectively. Quantitatively, DBLRGAN outperforms DBLRNet with slight advance (about 1\% improvement). As have mentioned above, the generator model in DBLRGAN aims to generate frames with similar pixel values as the sharp frames while the discriminator model along with the adversarial loss drives the generator to recover realistic images like real-word images. These two models complement each other and achieve better results. 
The results in Table \ref{table3} and \ref{table4} show that the improvement achieved by DBLRNet is more obvious than GAN model. While according to Figure \ref{figure5}, the deblurred frames generated by DBLRGAN are sharper than DBLRNett, \textit{e.g.}, the word "Bill" in the top row. $\alpha$ should be set as a little value because a bigger $\alpha$ will break the balance of content and adversarial loss, which causes worse performance of video deblurring.

\begin{table}
  \centering
    \caption{Performance comparisons with \cite{hyun2015generalized}, ~\cite{sellent2016stereo} and ~\cite{pan2017simultaneous} on the Blurred KITTI dataset in terms of the PSNR criterion. The best results are shown in bold, and the second best are underlined.}
    \begin{tabular}{c|cc}
    \toprule
    Methods & PSNR-LEFT & PSNR-RIGHT \\
    \hline
    Kim et al. & 28.25& 29.00\\
    Sellent et al. & 27.75& 28.52\\
    Pan et al. & \underline{30.24}& \underline{30.71}\\
    \hline
    DBLRNet (single) & 28.97 & 29.55\\
    DBLRNet (multi) & 29.94 & 30.33\\
    DBLRNet & 30.10 & 30.54 \\
    \hline
    DBLRGAN & \textbf{30.42} & \textbf{30.87} \\
    \bottomrule
    \end{tabular}
  \label{table4}
\end{table}

\begin{figure*} [tb]
\centering
\subfigure[]{
\begin{minipage}[b]{0.95\textwidth}
\includegraphics[width=0.95\linewidth]{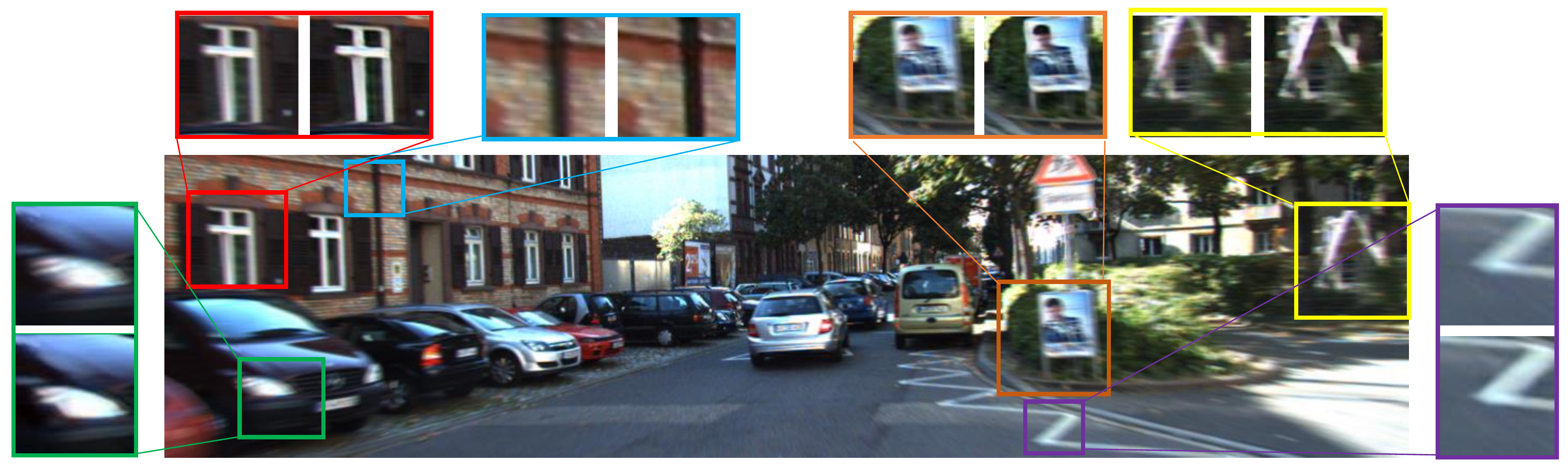}
\end{minipage}
}
\subfigure[]{
\begin{minipage}[b]{0.95\textwidth}
\includegraphics[width=0.95\linewidth]{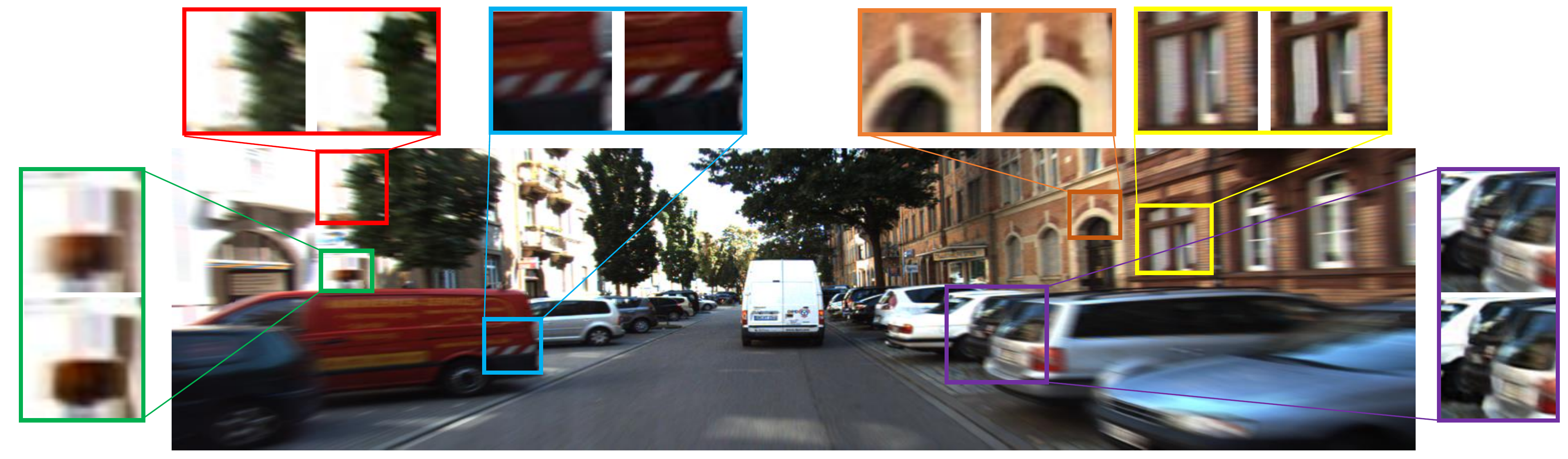}
\end{minipage}
}
\subfigure[]{
\begin{minipage}[b]{0.95\textwidth}
\includegraphics[width=0.95\linewidth]{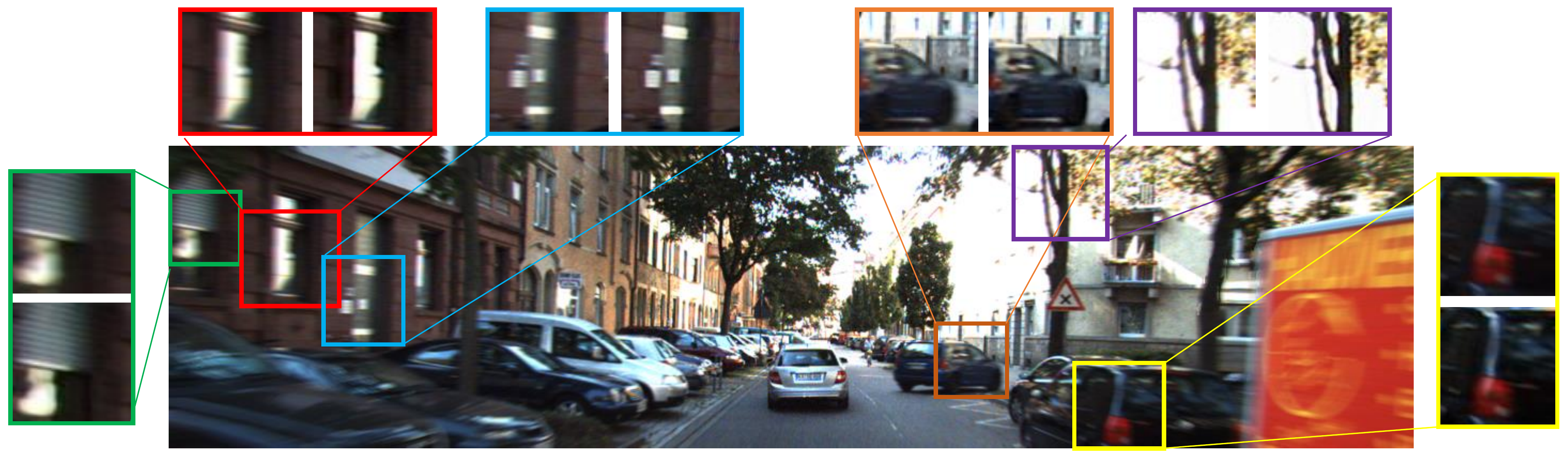}
\end{minipage}
}
\caption{Performance of our method on blurry videos caused by bokeh. The figure shows a sample frame from the Blurred KITTI dataset, which is captured from a car moving at a high speed. The blurs take place in the side area, while the center part is clear. We show a few pairs of zoomed-in patches from the frame before and after applying our method. The sharper edge demonstrates that our method can generalize well to other types of blurry videos.}
 \label{figure6}
\end{figure*}

Figure~\ref{figure4} and \ref{figure5} provide exemplar results on the quantitative and qualitative subsets of the VideoDeblurring dataset, respectively. Please notice the two columns corresponding to DBLRNet and DBLRGAN in Figure \ref{figure4}, especially the letters in the third row, where results of DBLRGAN are more photo-realistic than those of DBLRNet. The same case is observed in Figure \ref{figure5}. Letters in results of DBLRGAN are sharper than those of DBLRNet, which consistently shows that, DBLRGAN generates more realistic frames with finer textural details compared with DBLRNet.

All results of DBLRNet and DBLRGAN are obtained without aligning. Aligning images is computationally expensive and fragile \cite{su2016deep}. Kim et al. \cite{kim2017online} evaluate DBN model and find that the speed of DBN model without aligning is almost more than 20 times faster than it with aligning because aligning procedure is very time-consuming. Our proposed models enable the generation of high quality results without computing any alignment, which makes it highly efficient to scene types.

\subsection{Comparison with Existing Methods} 
To further verify the effectiveness of our models, we additionally compare the performance of DBLRNet and DBLRGAN with that of several state-of-the-art approaches on both the VideoDeblurring dataset and the KITTI dataset. 

\begin{figure} [tb]
	\centering
	\includegraphics[width=1\linewidth]{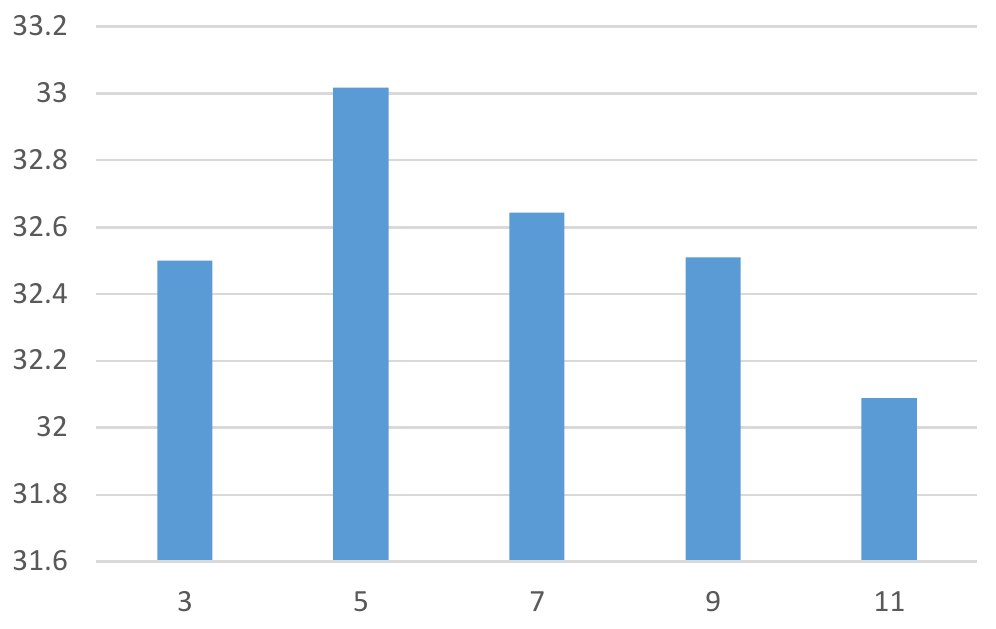}
	\caption{Performance comparisons of our method in terms of PSNR by varying the number of input frames.}
	\label{figure7}
\end{figure}

On the VideoDeblurring dataset, we compare our models with PSDEBLUR, WFA \cite{delbracio2015burst}, DBN \cite{su2016deep} and DBN (single). PSDEBLUR is the deblurred results of PHOTOSHOP. WFA is a method based on multiple frames as input. DBN achieves the state-of-the-art performance on the VideoDeblurring data set before this work. DBN (single) is a variant of DBN which stacks 5 copies of one single frame as input. Table \ref{table3} shows quantitative comparisons between our methods and these methods. Specially, the results indicate that our method significantly outperforms the DBN model by 3.14 db. Figure \ref{figure4} and \ref{figure5} also represent visual comparison between our models and these methods on both the quantitative (Figure \ref{figure4}) and qualitative (Figure \ref{figure5}) sub-datasets, respectively. Evidently our models achieves sharper results.

On the dataset of Blurred KITTI, we conduct comparison with \cite{hyun2015generalized}, \cite{sellent2016stereo} and \cite{pan2017simultaneous}. \cite{pan2017simultaneous} is a geometry based method and utilizes additional stereo information from image pairs. It is the current state-of-the-art on the Blurred KITTI dataset. We simply apply the DBLRNet trained on the VideoDeblurring dataset to the Blurred KITTI dataset and still achieve comparable results with \cite{pan2017simultaneous}. With the additional adversarial loss, DBLRGAN slightly outperforms \cite{pan2017simultaneous}. Please note that, our models are not specialized for the stereo setting.

\subsection{Different Frames \& Other Types of Blur}
\label{more_analysis}
\textbf{Different Frames.} We are curious about how the number of consecutive frames influences the performance of our DBLRGAN model. Thus we compare the PSNR values of the model by varying the number of input blurry frames. Making it more specific, on the VideoBlurring dataset, five kinds of settings, three, five, seven, nine and eleven continuous frames are taken as input to our model. Fig.~\ref{figure7} shows that our model with five frames as input achieves the best performance. With the increase of input frames, the PSNR values become lower. We suspect that, as our 3D convolution based network can extract powerful representations to describe short-term fast-varying motions occurring in continuous input frames, it is suitable to set the temporal span relatively small to capture the rapid dynamics across local adjacent frames.

\textbf{Generalize to Other Types of Blurry Videos.} Though our model is trained on the VideoDeblurring dataset, which includes only blurry frames caused by camera shakes, we are also curious about how it generalize to blurry videos of other blur types. To this end, we test it on videos from the Blurred KITTI dataset. Fig.~\ref{figure6} shows exemplar frames, which is captured by a camera mounted on a high-speed car. The dominated blur is cause by bokeh (see the comparison between the center area and the border area in the image), rather than camera shakes. As shown in the comparison of the enlarged patches, by applying our DBLRGAN model, the edges in the image become sharper. As discussed above, this verifies the advantage of our method capturing short-term fast-varying motions.

\textbf{Limitation.} Removing jumping artifacts is a challenge of video deblurring. As shown in Fig.~\ref{figure1} (col. 4\&5, row 2), there are also some jumping artifacts in the deblurred frames. Thus our method cannot solve it completely. However, the proposed model contributes to alleviate the unexpected temporal artifacts because it captures jointly spatial and temporal information encoded in neighboring frames. Even without post-processing and aligning, our proposed model can also achieve satisfied performance. Please refer to Fig.~\ref{figure4} and ~\ref{figure5}. Comparing with prior methods, when frames are severely blurred, our methods can generate better deblurred frames.

\section{Conclusions}
In this paper, we have resorted to spatio-temporal learning and adversarial training to recover sharp and realistic video frames for video deblurring. Specifically, we proposed two novel network models. The first one is our DBLRNet, which uses 3D convolutional kernels on the basis of deep residual neural networks. We demonstrated that DBLRNet is able to capture better spatio-temporal features, leading to improved blur removal. Our second contribution is DBLRGAN equipped with both the content loss and adversarial loss, which are complementary to each other, driving the model to generate visually realistic images. The experimental results on two standard benchmarks show that our proposed DBLRNet and DBLRGAN outperform the existing state-of-the-art methods in video deblurring.

\section*{Acknowledgment}
Kaihao Zhang's PhD scholarship is funded by Australian National University. Yiran Zhong's PhD scholarship is funded by CSIRO Data61. Hongdong Li is CI (Chief Inivestigator) on Australia Centre of Excellence for Robotic Vision (CE14) funded by Australia Research Council. This work is also supported by 2017 Tencent Rhino Bird Elite Graduate Program.

\ifCLASSOPTIONcaptionsoff
  \newpage
\fi



%



{
\bibliographystyle{ieeetr}
\bibliography{egbib_v1}
}

%


\begin{IEEEbiography}[{\includegraphics[width=1in,height=1.25in,clip,keepaspectratio]{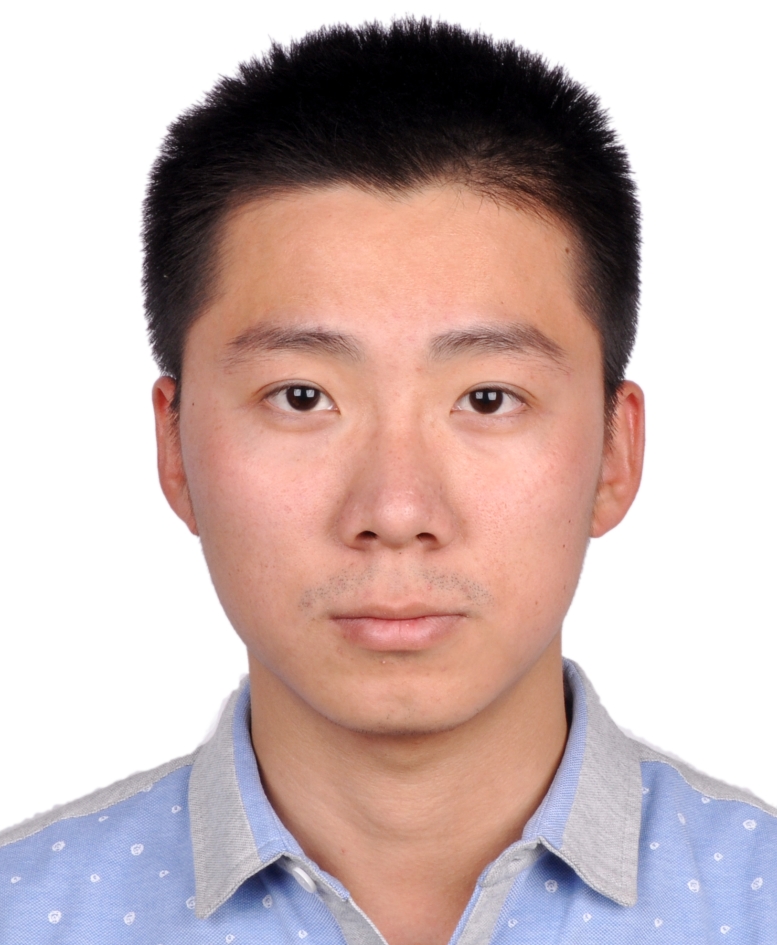}}]{Kaihao Zhang}
is currently pursuing the Ph.D. degree with the College of Engineering and Computer Science, The Australian National University, Canberra, ACT, Australia. Prior to that, he received the M.Eng. degree in computer application technology from the University of Electronic Science and Technology of China, Chengdu, China, in 2016. He worked at the Center for Research on Intelligent Perception and Computing, National Lab of Pattern Recognition, Institute of Automation, Chinese Academy of Sciences, Beijing, China for two years and the Tencent AI Laboratory, Shenzhen, China for one year. His research interests focus on video analysis and facial recognition with deep learning.
\end{IEEEbiography}

\begin{IEEEbiography}[{\includegraphics[width=1in,height=1.25in,clip,keepaspectratio]{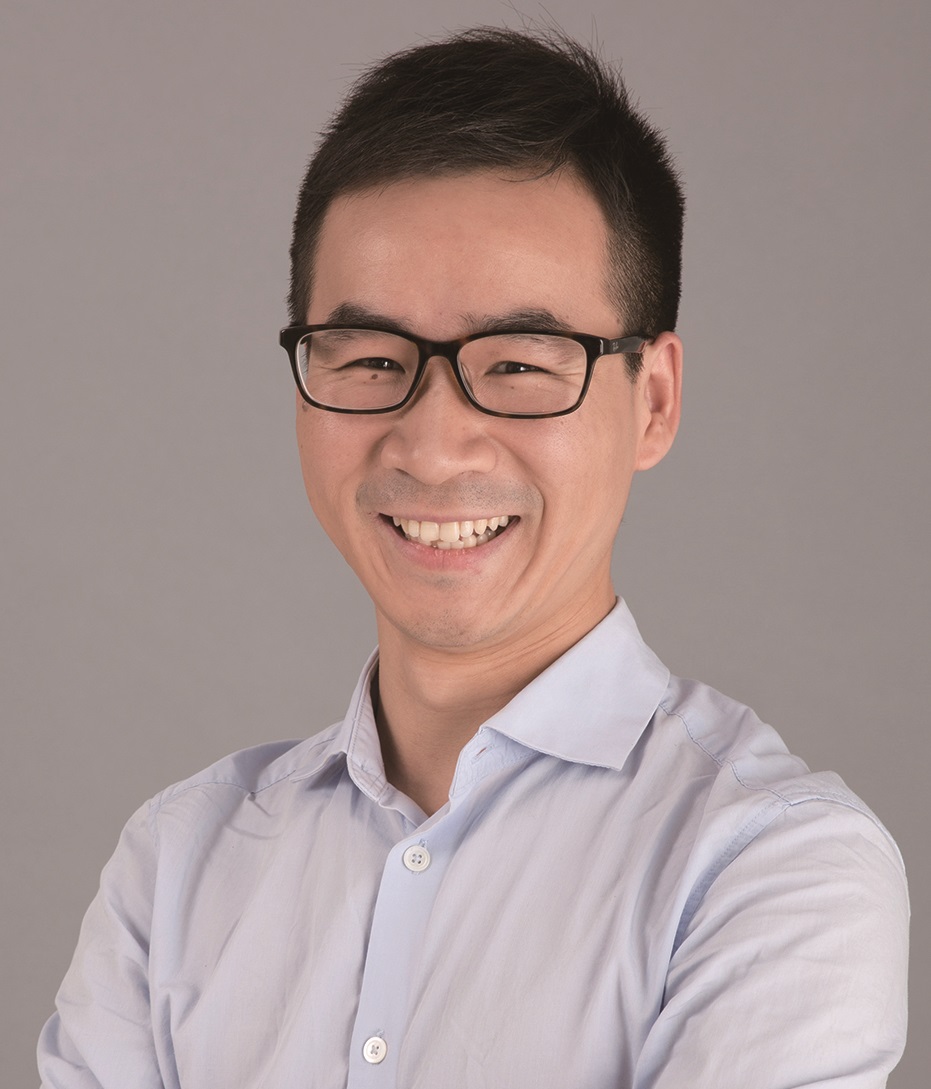}}]{Wenhan Luo}
is currently working as a senior researcher in the Tencent AI Lab, China. His research interests include several topics in computer vision and machine learning, such as motion analysis (especially object tracking), image/video quality restoration, reinforcement learning. Before joining Tencent, he received the Ph.D. degree from Imperial College London, UK, 2016, M.E. degree from Institute of Automation, Chinese Academy of Sciences, China, 2012 and B.E. degree from Huazhong University of Science and Technology, China, 2009.
\end{IEEEbiography}
-
\begin{IEEEbiography}[{\includegraphics[width=1in,height=1.25in,clip,keepaspectratio]{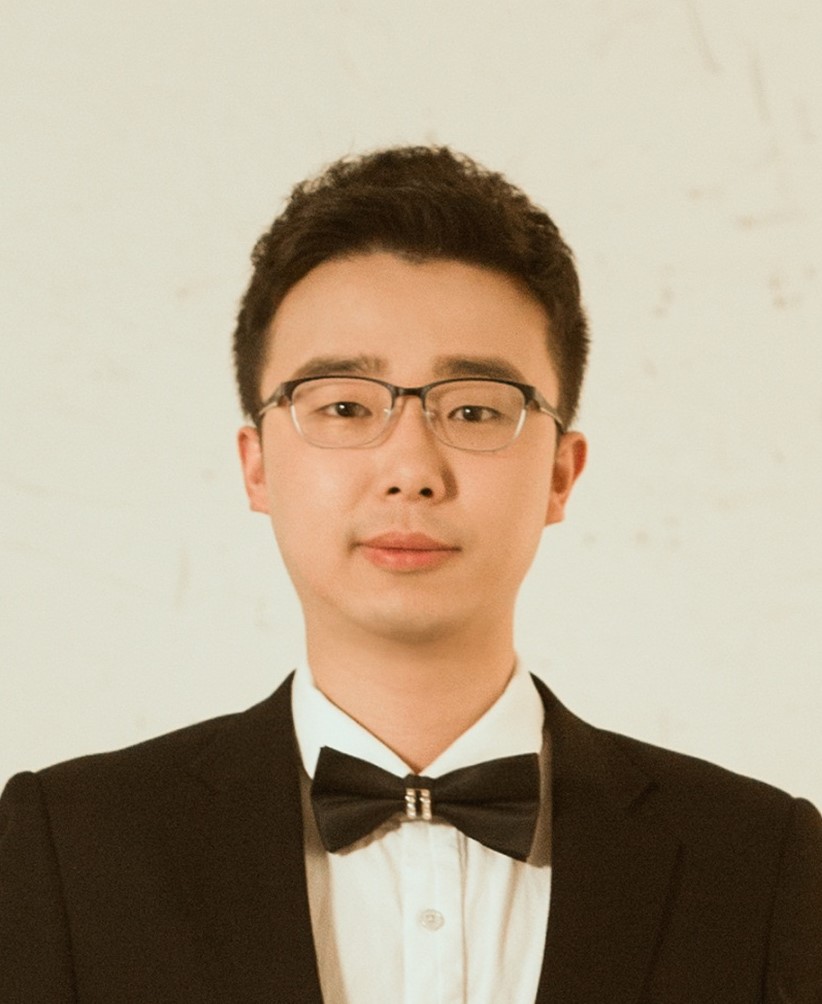}}]{Yiran Zhong} received the M.Eng in information and electronics engineering in 2014 with the first class honor from The Australian National University, Canberra, Australia. After two years of research assistant, he becomes a PhD student in the College of Engineering and Computer Science, The Australian National University, Canberra, Australia and Data61, CSIRO, Canberra, Australia. He won the ICIP Best Student Paper Award in 2014. His current research interests include geometric computer vision, machine learning and deep learning.

\end{IEEEbiography}

\begin{IEEEbiography}[{\includegraphics[width=1in,height=1.25in,clip,keepaspectratio]{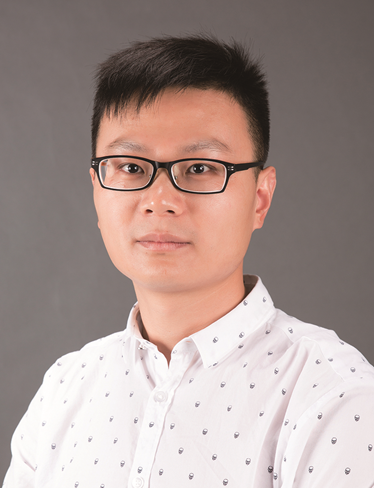}}]{Lin Ma}(M’13) 
received the B.E. and M.E. degrees in computer science from the Harbin Institute of Technology, Harbin, China, in 2006 and 2008, respectively, and the Ph.D. degree from the Department of Electronic Engineering, The Chinese University of Hong Kong, in 2013. He was a Researcher with the Huawei Noah’Ark Laboratory, Hong Kong, from 2013 to 2016. He is currently a Principal Researcher with the Tencent AI Laboratory, Shenzhen, China. His current research interests lie in the areas of computer vision, multimodal deep learning, specifically for image and language, image/video understanding, and quality assessment. 

Dr. Ma received the Best Paper Award from the Pacific-Rim Conference on Multimedia in 2008. He was a recipient of the Microsoft Research Asia Fellowship in 2011. He was a finalist in HKIS Young Scientist Award in engineering science in 2012.

\end{IEEEbiography}

\begin{IEEEbiography}[{\includegraphics[width=1in,height=1.25in,clip,keepaspectratio]{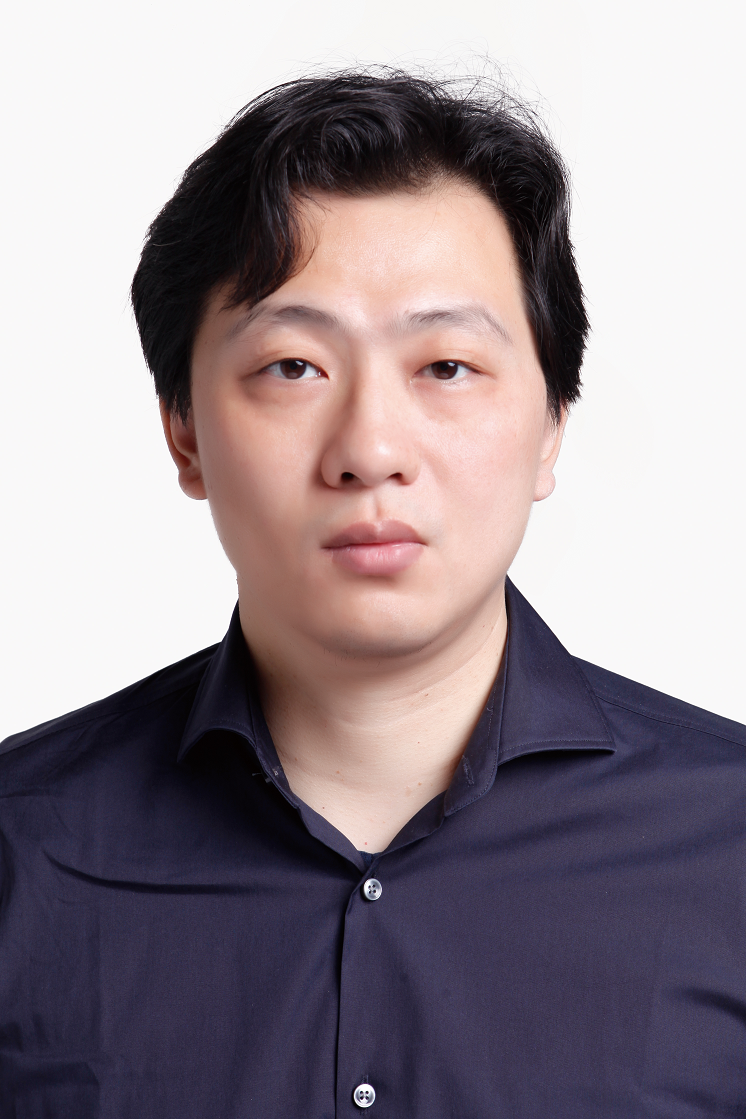}}]{Wei Liu}
is currently a Distinguished Scientist of Tencent AI Lab and a director of Computer Vision Center. Prior to that, he received the Ph.D. degree in EECS from Columbia University, New York, NY, USA, and was a research scientist of IBM T. J. Watson Research Center, Yorktown Heights, NY, USA. Dr. Liu has long been devoted to research and development in the fields of machine learning, computer vision, information retrieval, big data, etc. Till now, he has published more than 150 peer-reviewed journal and conference papers, including Proceedings of the IEEE, TPAMI, TKDE, IJCV, NIPS, ICML, CVPR, ICCV, ECCV, KDD, SIGIR, SIGCHI, WWW, IJCAI, AAAI, etc. His research works win a number of awards and honors, such as the 2011 Facebook Fellowship, the 2013 Jury Award for best thesis of Columbia University, the 2016 and 2017 SIGIR Best Paper Award Honorable Mentions, and the 2018 "AI's 10 To Watch" honor. Dr. Liu currently serves as an Associate Editor to several international leading AI journals and an Area Chair to several international top-tier AI conferences, respectively.

\end{IEEEbiography}

\begin{IEEEbiography}[{\includegraphics[width=1in,height=1.25in,clip,keepaspectratio]{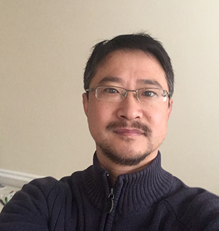}}]{Hongdong Li} is currently a Reader with the Computer Vision Group of ANU (Australian National University). He is also a Chief Investigator for the Australia ARC Centre of Excellence for Robotic Vision (ACRV). His research interests include 3D vision reconstruction, structure from motion, multi-view geometry, as well as applications of optimization methods in computer vision.  Prior to 2010, he was with NICTA Canberra Labs working on the “Australia Bionic Eyes” project.  He is an Associate Editor for IEEE T-PAMI, and served as Area Chair in recent year ICCV, ECCV and CVPR.  He was a Program Chair for ACRA 2015 - Australia Conference on Robotics and Automation, and a Program Co-Chair for ACCV 2018 - Asian Conference on Computer Vision. He won a number of prestigious best paper awards in computer vision and pattern recognition, and was the receipt for the CVPR Best Paper Award in 2012 and the Marr Prize Honorable Mention in 2017. 

\end{IEEEbiography}





\end{document}